\documentclass{article}
\usepackage{ijcai24}

\usepackage{times}
\usepackage{soul}
\usepackage{url}
\usepackage[hidelinks]{hyperref}
\usepackage[utf8]{inputenc}
\usepackage[small]{caption}
\usepackage{graphicx}
\usepackage{amsmath}
\usepackage{amsthm}
\usepackage{booktabs}
\usepackage{algorithm}
\usepackage{algorithmic}
\usepackage[switch]{lineno}

\usepackage{multicol}
\usepackage{multirow}
\usepackage{subfig}
\usepackage{amssymb}
\usepackage{color}
\usepackage{xcolor}

\newtheorem{example}{Example}
\newtheorem{theorem}{Theorem}
\newtheorem{assumption}{Assumption}
\newtheorem{definition}{Definition}
\newtheorem{corollary}{Corollary}
\newtheorem{remark}{Remark}

\title{Subgraph Pooling: Tackling Negative Transfer on Graphs}

\author{
Zehong Wang$^1$\and
Zheyuan Zhang$^1$\and
Chuxu Zhang$^2$\thanks{Corresponding Authors.}\and
Yanfang Ye$^1$\footnotemark[1] \\
\affiliations
$^1$University of Notre Dame, Indiana, USA\\
$^2$Brandeis University, Massachusetts, USA\\
\emails
\{zwang43, zzhang42, yye7\}@nd.edu,
chuxuzhang@brandeis.edu
}
\begin{document}

\maketitle

\begin{abstract}
Transfer learning aims to enhance performance on a target task by using knowledge from related tasks. However, when the source and target tasks are not closely aligned, it can lead to reduced performance, known as negative transfer. Unlike in image or text data, we find that negative transfer could commonly occur in graph-structured data, even when source and target graphs have semantic similarities. Specifically, we identify that structural differences significantly amplify the dissimilarities in the node embeddings across graphs. To mitigate this, we bring a new insight in this paper: for semantically similar graphs, although structural differences lead to significant distribution shift in node embeddings, their impact on subgraph embeddings could be marginal. Building on this insight, we introduce Subgraph Pooling (SP) by aggregating nodes sampled from a k-hop neighborhood and Subgraph Pooling++ (SP++) by a random walk, to mitigate the impact of graph structural differences on knowledge transfer. We theoretically analyze the role of SP in reducing graph discrepancy and conduct extensive experiments to evaluate its superiority under various settings. The proposed SP methods are effective yet elegant, which can be easily applied on top of any backbone Graph Neural Networks (GNNs). Our code and data are available at: \url{https://github.com/Zehong-Wang/Subgraph-Pooling}.

%Transfer learning aims to boost the learning on the target task leveraging knowledge learned from other relevant tasks. However, when the source and target are not closely related, the learning performance may be adversely affected, a phenomenon known as negative transfer. In this paper, we investigate the negative transfer in graph transfer learning, which is important yet underexplored. We reveal that, unlike image or text, negative transfer commonly occurs in graph-structured data, even when source and target graphs share semantic similarities. Specifically, we identify that structural differences significantly amplify the dissimilarities in the node embeddings across graphs. To mitigate this, we bring a new insight: for semantically similar graphs, although structural differences lead to significant distribution shift in node embeddings, their impact on subgraph embeddings could be marginal. Building on this insight, we introduce two effective yet elegant methods, Subgraph Pooling (SP) and Subgraph Pooling++ (SP++), that transfer subgraph-level knowledge across graphs. We theoretically analyze the role of SP in reducing graph discrepancy and conduct extensive experiments to evaluate its superiority under various settings. Our code and datasets are available at: \url{https://github.com/Zehong-Wang/Subgraph-Pooling}\footnote{Extended version in \url{https://arxiv.org/abs/2402.08907}.}.

\end{abstract}

\section{Introduction}

Graph Neural Networks (GNNs) are widely employed for graph mining tasks across various fields~\cite{gaudelet2021utilizing,kipf2017semisupervised,he2020lightgcn}. Despite their remarkable success in graph-structured datasets, these methods exhibit limitations in label sparse scenarios~\cite{dai2022towards}. This restricts the applications of GNNs in real-world datasets where label acquisition is challenging or impractical. To address the issue, transfer learning~\cite{zhuang2020comprehensive} emerges as a solution, which aims to transfer knowledge from a label-rich source graph to a label-sparse target graph through fine-tuning or prompting.

However, the success of transfer learning is not always guaranteed~\cite{wang2019characterizing,zhang2022survey}. If the source and target tasks are not closely aligned, transferring knowledge from such weakly related sources may impair the performance on the target, known as negative transfer~\cite{wang2019characterizing}. By interpreting transfer learning as a generalization problem, \cite{wang2019characterizing} demonstrated that negative transfer derives from the divergence between joint distributions of the source and target tasks. To this end, researchers employed adversarial learning~\cite{wu2020unsupervised}, causal learning~\cite{chen2022learning}, or domain regularizer~\cite{you2023graph} to develop domain-invariant encoders, reducing the distribution shift between the source and target.

\begin{figure*}[!h]
    \centering
    \subfloat{\includegraphics[width=0.25\linewidth]{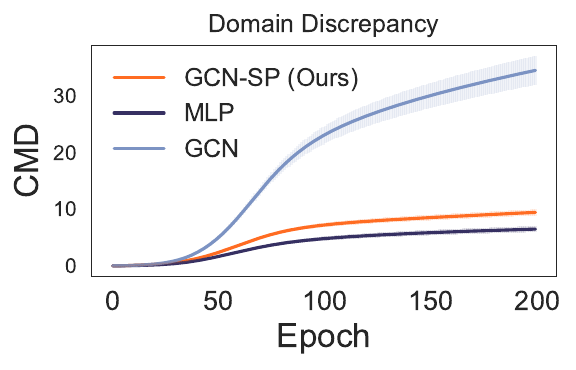}}
    \subfloat{\includegraphics[width=0.25\linewidth]{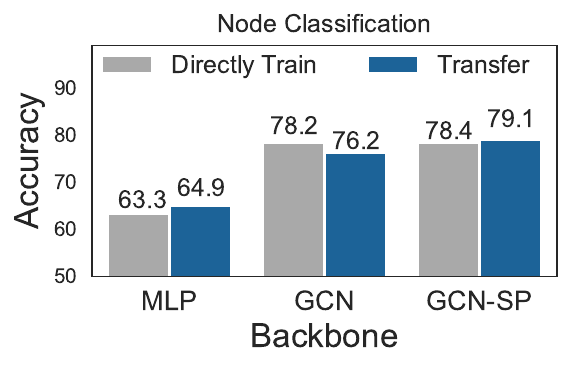}}
    \subfloat{\includegraphics[width=0.25\linewidth]{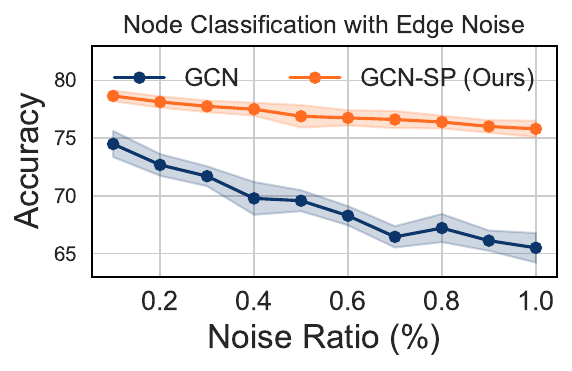}}
    \subfloat{\includegraphics[width=0.25\linewidth]{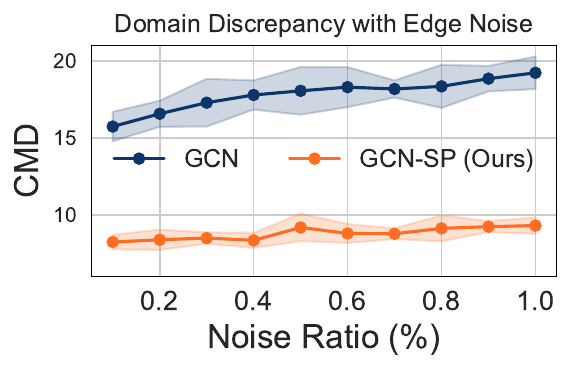}}
    \caption{Structural differences between the source (\texttt{DBLP}) and target (\texttt{ACM}) amplify the distribution shift on nodes embeddings. \textbf{Left:} We illustrate the discrepancy (CMD value) between node embeddings of the source and target during pre-training, and compare the performance of direct training on the target (gray) and transferring knowledge from the source to the target (blue). A large discrepancy results in negative transfer. \textbf{Right:} We introduce structural noise in the target graph through random edge permutation. Even minor permutations can enlarge the discrepancy (and thus aggravate negative transfer) in vanilla GCN, yet our method effectively mitigates the issue.}
    \label{fig:negative transfer}
\end{figure*}

In this paper, we systematically analyze the negative transfer on graphs, a lack of existing works. We find negative transfer often occurs in graph datasets even when the source and target graphs are semantically similar. The observation contrasts with visual and textual modalities, where similar sources typically enhance the performance on targets~\cite{zhuang2020comprehensive}. We identify that the issue stems from the graph structural differences between the source and target graphs, which leads to significant distribution shifts on node embeddings. For example, in financial transaction networks \cite{weber2019anti} collected over different time intervals, the patterns of transactions can vary significantly due to the influence of social events or policy changes. These evolving patterns notably alter the local structure of users, leading to divergence in user embedding. To tackle the negative transfer on graphs, we introduce two straightforward yet effective methods called Subgraph Pooling (SP) and Subgraph Pooling++ (SP++) to reduce the discrepancy between graphs.
Our major contributions are summarized as follows:
\begin{itemize}
    \item \textbf{Negative Transfer in GNNs.} We systematically analyze the negative transfer on graphs. We find that the structural difference between the source and target graphs amplifies distribution shifts on node embeddings, as the aggregation process of GNNs is highly sensitive to perturbations in graph structures. To address this issue, we present a novel insight: for semantically similar graphs, although structural differences lead to significant distribution shift in node embeddings, their impact on subgraph embeddings could be marginal.

    \item \textbf{Subgraph Pooling to Tackle Negative Transfer.} Building upon this insight, we introduce plug-and-play modules Subgraph Pooling (SP) and Subgraph Pooling++ (SP++) to mitigate the negative transfer. The key idea is to transfer subgraph information across source and target graphs to prevent the distribution shift. Notably, we provide a comprehensive theoretical analysis to clarify the mechanisms behind Subgraph Pooling.

    \item \textbf{Generality and Effectiveness.} Subgraph Pooling is straightforward to implement and introduces no additional parameters. It involves simple sampling and pooling operations, making it easily applicable to any GNN backbone. We conduct extensive experiments to demonstrate that our method can significantly surpass existing baselines under multiple transfer learning settings.
\end{itemize}

\section{Negative Transfer in GNNs}

% Preliminary of graphs

\subsection{Preliminary of Graph Transfer Learning}

% Preliminary of semi-supervised transfer learning. 

Semi-supervised graph learning is a common setting in real-world applications~\cite{kipf2017semisupervised}. In this work, we study negative transfer in semi-supervised graph transfer learning for node classification, while our analysis is also applicable for other transfer learning settings~\cite{wu2020unsupervised}. Semi-supervised transfer learning focuses on transferring knowledge from a label-rich source $\mathcal{D}_S$ to a label-sparse target $\mathcal{D}_T$. We represent the joint distribution over the source and target as $P_S(\mathcal{X}, \mathcal{Y})$ and $P_T(\mathcal{X}, \mathcal{Y})$, respectively, where $\mathcal{X}$ indicates the random input space and $\mathcal{Y}$ is the output space. The labeled training instances are sampled as $\mathcal{D}_S = \{(x_i^s, y_i^s)\}_{i=1}^{n_s} \sim P_S(\mathcal{X}, \mathcal{Y})$ and $\mathcal{D}_T^L = \{(x_i^t, y_i^t)\}_{i=1}^{n_t^l} \sim P_T(\mathcal{X}, \mathcal{Y})$, while the unlabeled instances are $\mathcal{D}^U_T = \{(x_i^t)\}_{i=1}^{n_t^u} \sim P_T(\mathcal{X})$, combining to form $\mathcal{D}_T = (\mathcal{D}_T^L, \mathcal{D}_T^U)$. The objective is to develop a hypothesis function $h: \mathcal{X} \to \mathcal{Y}$ that minimizes the empirical risk on the target $R_T(h) = \mathbf{Pr}_{(x, y) \sim \mathcal{D}_T} (h(x) \neq y)$.

% Preliminary of semi-supervised graph transfer learning

Considering graph-structured data, a graph is represented as $\mathcal{G} = (\mathcal{V}, \mathcal{E})$, where $\mathcal{V} = \{v_1, ..., v_n\}$ is the node set and $\mathcal{E} \subseteq \mathcal{V} \times \mathcal{V}$ is the edge set. Each node $i \in \mathcal{V}$ is associated with node attributes $\mathbf{x}_i \in \mathbb{R}^{d}$ and a class $y_i \in \{1, ..., C\}$, with $C$ being the total number of classes. Additionally, each graph has an adjacency matrix $\mathbf{A} \in \{0, 1\}^{n \times n}$, where $\mathbf{A}_{ij} = 1$ iff $(i, j) \in \mathcal{E}$, otherwise $\mathbf{A}_{ij} = 0$. In the semi-supervised transfer setting, we have the source graph $\mathcal{G}^s = (\mathcal{V}^s, \mathcal{E}^s)$ and target graph $\mathcal{G}^t = (\mathcal{V}^t, \mathcal{E}^t)$. For simplicity, we assume these graphs share the same feature space $\mathbf{X}^s \in \mathbb{R}^{n_s \times d}$ and $\mathbf{X}^t \in \mathbb{R}^{n_t \times d}$, as well as a common label space $y^s, y^t \in \{1, ..., C\}$. We employ a GNN backbone $f(\cdot)$ to encode nodes into embeddings $\mathbf{Z}^s, \mathbf{Z}^t$ and then use a classifier $g(\cdot)$ for predictions. The joint distributions over the source and target graphs are $P_S(\mathcal{Z}, \mathcal{Y})$ and $P_T(\mathcal{Z}, \mathcal{Y})$, where $\mathcal{Z}$ denotes the node embedding space.

% Definition of semi-supervised graph transfer learning

\begin{definition}[Semi-supervised Graph Transfer Learning]
    The aim is to transfer knowledge from a label-rich source graph $\mathcal{G}^s$ to a semantically similar label-sparse target graph $\mathcal{G}^t$ for enhancing node classification performance. The joint distributions $P(\mathcal{Z}, \mathcal{Y})$ over the source and target are different, where $P_S(\mathcal{Y} | \mathcal{Z}) = P_T(\mathcal{Y} | \mathcal{Z})$ and $P_S(\mathcal{Z}) \neq P_T(\mathcal{Z})$.
\end{definition}

% How to measure the distribution shift
%   Definition of CMD

While the conditional distributions $P(\mathcal{Y} | \mathcal{Z})$ over the source and target are identical, their marginal distributions $P(\mathcal{Z})$ are different. To quantify this discrepancy, researchers utilize metrics such as Maximum Mean Discrepancy (MMD)~\cite{gretton2006kernel}, Center Moment Discrepancy (CMD)~\cite{zellinger2017central}, or Wasserstein Distance~\cite{zhu2023explaining}, to measure node similarities in complex spaces. We use CMD due to its computational efficiency:
\begin{align}
    \nonumber d_{CMD} & = \frac{1}{| n_s - n_t |} \left\| \mathbb{E}(Z^s) - \mathbb{E}(Z^t) \right\|_2   \\
                      & + \sum_{k=2}^K \frac{1}{| n_s - n_t |^k} \left\| c_k(Z^s) - c_k(Z^t) \right\|_2,
\end{align}
where $c_k(\cdot)$ denotes the $k$-th order central moment (with $K=3$). A high CMD value indicates a considerable shift in marginal distributions between the source and target. This shift essentially results in a divergence between joint distributions $P_S(\mathcal{Z}, \mathcal{Y})$ and $P_T(\mathcal{Z}, \mathcal{Y})$, which may hinder or even degrade performance on the target~\cite{wang2019characterizing}.

\subsection{Why Negative Transfer Happens?}

% One common issue in graph transfer learning. 
%   Structure information causes large discrepancy, thus leading to server negative transfer. 
%   The observation of graph discrepancy and transfer learning performance

Negative transfer occurs in GNNs even if the source and target graphs are semantically similar. This issue is attributed to the sensitivity of GNNs on graph structures, where different structures diverge marginal distributions $P_S(\mathcal{Z})$ and $P_T(\mathcal{Z})$ between the source and target graphs. To support this claim, we illustrate the impact of structural differences on distribution discrepancy (CMD value) and transfer learning performance in Figure \ref{fig:negative transfer} \textbf{(Left)}. 
Particularly, the discrepancy remains low if graph structure is ignored (using MLP), ensuring the performance gain of transfer learning. Conversely, incorporating structural information through GNNs can increase the discrepancy, resulting in negative transfer.

% Further analyze the impact of graph structure
%   The observation of the impact of noisy graph structure

Based on the observations, we consider that GNNs may project semantically similar graphs into distinct spaces, unless their structures are very similar. To further reveal the phenomenon, we delve into the aggregation process of GNNs. For any GNN architecture, each node is associated with a computational tree, through which messages are passed and aggregated from leaves to the root. Only closely aligned structures can lead to similar computational tree distributions across graphs, thereby ensuring closely matched node embeddings. However, this requirement is often impractical in many graph datasets. Even minor perturbations in the graph structure can dramatically alter the computational tree, either by dropping critical branches or by introducing noisy connections. Furthermore, a single perturbation can impact the computational trees of multiple nodes, thus altering the computational tree distributions across the graph. We demonstrate the impact of structure perturbations in Figure \ref{fig:negative transfer} \textbf{(Right)}. In conclusion, structural differences between the source and target result in distinct computational tree distributions, culminating in a significant distribution shift in node embeddings.

\subsection{Analyzing The Impact of Structure}

The above analysis suggests that mitigating the impact of graph structure on node embeddings is crucial for alleviating negative transfer. Existing works implicitly or explicitly handle this issue. For instance, some researchers utilize adversarial learning~\cite{wu2020unsupervised,dai2022graph} or domain regularizers~\cite{you2023graph} to develop domain-invariant GNN encoders, which consistently project graphs with different structures into a unified embedding space. However, these methods lack generalizability to new, unseen graphs and are sensitive to structural perturbations. Alternatively, another line of work employs causal learning~\cite{wu2022handling,chen2022learning} or augmentation~\cite{liu2022local} to train encoders robust to structural distribution shift. Yet, these methods essentially generate additional training graphs to enhance robustness against minor structural perturbations, instead of considering the fundamental nature of graph structures. Unlike these two approaches, we present a novel insight to solve the issue: for semantically similar graphs, although structural differences lead to significant distribution shift in node embeddings, their impact on subgraph embeddings could be marginal. To better describe this phenomenon, we introduce node-level and subgraph-level discrepancy as metrics to evaluate the influence of graph structures.

\begin{definition}[Node-level Discrepancy]
    For nodes $u \in \mathcal{V}^s$ in source graph and $v \in \mathcal{V}^t$ in target graph, we have
    \begin{equation}
        \mathbb{E}_{u \in \mathcal{V}^s, v \in \mathcal{V}^t} \frac{\mathbf{z}_u^T \mathbf{z}_u}{\mathbf{z}_u^T \mathbf{z}_v} \ge \lambda,
    \end{equation}
    where $\lambda$ denotes the node-level discrepancy.
    \label{def:node discrepancy}
\end{definition}

\begin{definition}[Subgraph-level Discrepancy]
    For node $u \in \mathcal{V}^s$ with surrounding subgraph $\mathcal{S}^s_u = (\mathcal{V}^s_u, \mathcal{E}^s_u)$ and node $v \in \mathcal{V}^t$ with surrounding subgraph $\mathcal{S}^t_v = (\mathcal{V}^t_v, \mathcal{E}^t_v)$, we have
    \begin{equation}
        \mathbb{E}_{u \in \mathcal{V}^s, v \in \mathcal{V}^t} \left\| \frac{1}{n^s_u + 1} \sum_{i \in \mathcal{V}^s_u} \mathbf{z}_i - \frac{1}{m^t_v + 1} \sum_{j \in \mathcal{V}^t_v} \mathbf{z}_j \right\| \le \epsilon
    \end{equation}
    where $n^s_u = | \mathcal{V}^s_u |$, $m^t_v = | \mathcal{V}^t_v |$, and $\epsilon$ denotes the subgraph-level discrepancy.
    \label{def:subgraph discrepancy}
\end{definition}

% Explain subgraph-level discrepancy and link to the proposed method.

\begin{table}[!t]
    \centering
    \resizebox{\linewidth}{!}{
        \begin{tabular}{lcccc}
            \toprule
                                 & ACM $\to$ DBLP & DBLP $\to$ ACM & Arxiv T1 & Arxiv T3 \\ \midrule
            $\lambda$            & 2.413          & 2.353          & 2.134    & 2.683    \\
            $\epsilon$ ($k$-hop) & 0.212          & 0.380          & 0.191    & 0.203    \\
            $\epsilon$ (RW)      & 0.166          & 0.322          & 0.184    & 0.212    \\ \bottomrule
        \end{tabular}
    }
    \caption{Although node-level discrepancy ($\lambda$) between source and target is high, the subgraph-level discrepancy ($\epsilon$) remains low. $k$-hop and RW (Random Walk) indicate two subgraph sampling methods. }
    \label{tab:discrepancy}
\end{table}

Intuitively, a high $\lambda$ value suggests a significant distinction in node embeddings between the source and target, which potentially leads to negative transfer. On the other hand, a low value of $\epsilon$ indicates similar subgraph embeddings across the source and target, which potentially prevents negative transfer. We demonstrate the impact of graph structures on these two measurements using real-world datasets, as detailed in Table \ref{tab:discrepancy}. Although the node embeddings are distinct between the source and target owing to the impact of structural differences (as indicated by high $\lambda$), the subgraph embeddings remain similar across graphs (as indicated by low $\epsilon$). Drawing on these insights, we propose to directly transfer the subgraph information across graphs to enhance transfer learning performance by mitigating the impact of graph structures.

\section{Overcoming Negative Transfer}

\subsection{Subgraph Pooling}

\begin{figure*}[!t]
    \centering
    \subfloat{\includegraphics[width=0.25\linewidth]{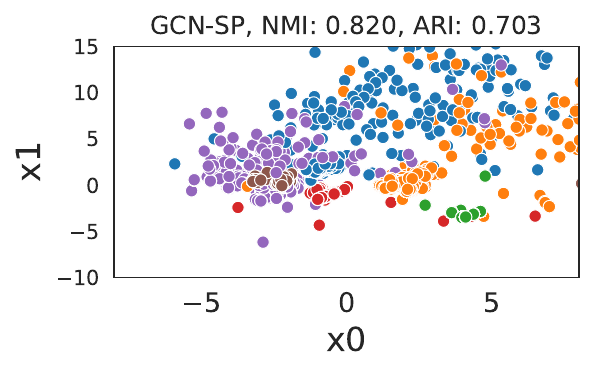}}
    \subfloat{\includegraphics[width=0.25\linewidth]{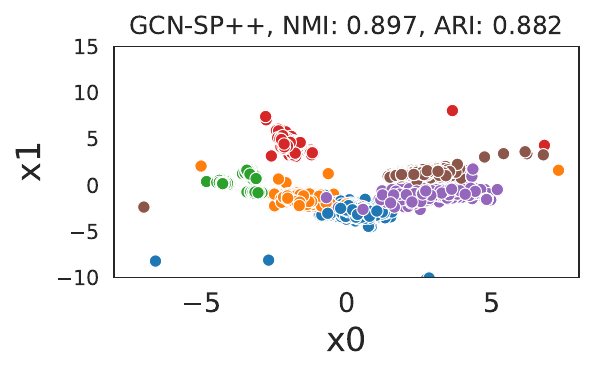}}
    \subfloat{\includegraphics[width=0.25\linewidth]{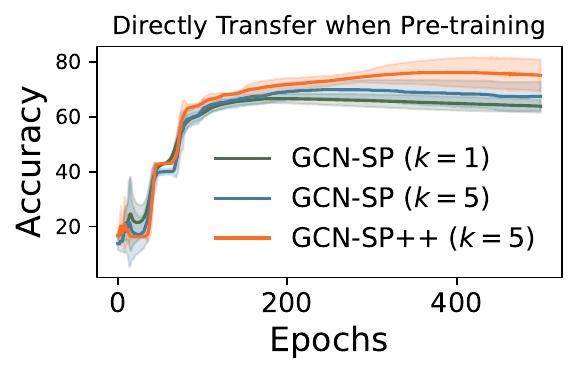}}
    \subfloat{\includegraphics[width=0.25\linewidth]{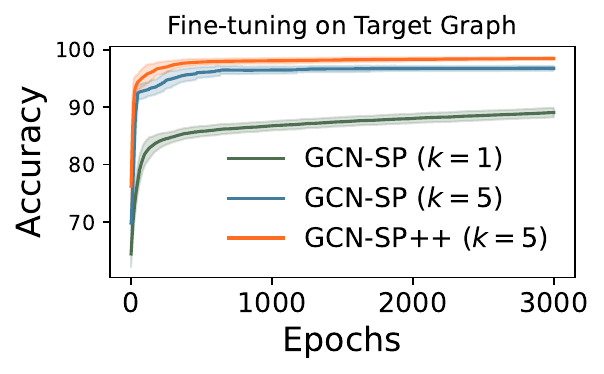}}
    \caption{Subgraph Pooling++ (SP++) mitigates the risk of over-smoothing derived from a large pooling kernel. We conduct transfer learning from \texttt{ACM} to \texttt{DBLP}. \textbf{Left:} Illustration of the subgraph embeddings on the target graph with $k=5$, where SP++ leveraging RW sampler has a clearer boundary. \textbf{Right:} Transfer learning performance during pre-training and fine-tuning, where SP++ achieves better.}
    \label{fig:rw sampling}
\end{figure*}

The goal of node-level graph transfer learning is to minimize the empirical risk (loss) on the target distribution: 
\begin{equation}
    \min \mathbb{E}_{(z, y) \sim P_T(\mathcal{Z}, \mathcal{Y})} [\mathcal{L}(g(z), y)],
    \label{eq:final objective}
\end{equation}
where $P_T(\mathcal{Z}, \mathcal{Y})$ is the joint distribution over the target graph, $z$ denotes the node embeddings encoded by a GNN backbone $f_{\Theta}(\cdot)$ with parameters $\Theta$, and $g(\cdot)$ is a linear classifier:
\begin{equation}
    \hat{\mathbf{Y}} = g(\mathbf{Z}), \mathbf{Z} = f(\mathbf{A}, \mathbf{X}, \Theta).
\end{equation}
However, owing to the scarcity of labels, we cannot exactly describe the joint distribution $P_T(\mathcal{Z}, \mathcal{Y})$~\cite{wenzel2022assaying}. Consequently, directly optimizing Eq. \ref{eq:final objective} on the target graph may lead to overfitting~\cite{mallinar2022benign}. Additionally, based on the above discussion, it is non-trivial to obtain a suitable initialization by pre-training the model on a semantically similar, label-rich source graph, as structural differences enlarge the discrepancy between the marginal distributions $P_S(\mathcal{Z})$ and $P_T(\mathcal{Z})$. To overcome the limitation, we introduce Subgraph Pooling (SP), a plug-and-play method that leverages subgraph information to diminish the discrepancy between the source and target. This approach is based on the following assumption.

\begin{assumption}
    For two semantically similar graphs, the subgraph-level discrepancy $\epsilon$ is small enough.
    \label{assump:subgraph discrepancy}
\end{assumption}
\begin{remark}
    We empirically validate the assumption in Table \ref{tab:discrepancy} and consider it matches many real-world graphs. For example, considering papers from two citation networks, if they share the same research field, they tend to have similar local structures, e.g., neighbors, since papers within the same domain often reference a core set of foundational works. Additionally, this pattern extends to social networks, where individuals with similar interests or professional backgrounds are likely to have comparable connection patterns, reflecting shared community norms or communication channels.
\end{remark}

% In the original GCN, the graph aggregation function at $k$-th order is defined as
% \begin{equation}
%     \mathbf{Z}^{(k)} = \sigma(\hat{\mathbf{D}}^{-1/2} \hat{\mathbf{A}} \hat{\mathbf{D}}^{-1/2} \mathbf{Z}^{(k-1)} \Theta^{(k)}), 
% \end{equation}
% where $\sigma$ is the activation, $\mathbf{Z} = \mathbf{Z}^{(K)}$ is the result of the last layer, and $\mathbf{Z}^{(0)} = \mathbf{X}$. The $\hat{\mathbf{A}} = \mathbf{A} + \mathbf{I}$ is the optimized adjacency matrix and $\hat{\mathbf{D}}$ denotes the corresponding diagonal degree matrix. The $\mathbf{Z} \in \mathbb{R}^{n \times d}$ is the learned node embeddings. 

The key idea of Subgraph Pooling is to transfer subgraph-level knowledge across graphs. The method is applicable for arbitrary GNNs by adding a subgraph pooling layer at the end of backbone. Specifically, in the SP layer, we first sample the subgraphs around nodes and then perform pooling to generate subgraph embeddings for each node. The choice of sampling and pooling functions can be arbitrary. Here we consider a straightforward sampling method, defined as the $k$-hop subgraph around each node:
\begin{equation}
    \mathcal{N}_s(i) = \text{Sample}_{k\text{-hop}}(\mathcal{G}, i).
\end{equation}
Subsequently, we pool the subgraph for each node:
\begin{equation}
    \mathbf{h}_i = \frac{1}{|\mathcal{N}_s(i)| + 1} \sum_{j \in \mathcal{N}_s(i) \cup i} w_{ij} \mathbf{z}_j.
\end{equation}
where $\mathbf{h}_i \in \mathbf{H}$ represents the subgraph embeddings (the new embeddings for each node), utilized in training the classifier $g(\cdot)$. $w_{ij}$ denotes the pooling weight, which can be either learnable or fixed. Empirically, the MEAN pooling function is effective enough to achieve desirable transfer performance. By leveraging subgraph information, the SP layer reduces the discrepancy (CMD value) between node embeddings in the source and target, thereby enhancing transfer learning performance, illustrated in Figure \ref{fig:negative transfer} \textbf{(Left)}. Furthermore, this also reduces the sensitivity on structural perturbations, as evidenced in Figure \ref{fig:negative transfer} \textbf{(Right)}. 

The integration of the SP layer into GNN architectures does not substantially increase time complexity. We have three-fold considerations. Firstly, the SP layer functions as a non-parametric GNN layer, thus imposing no additional burden on model optimization and enjoying the computational efficiency of existing GNN libraries~\cite{fey2019fast}. Secondly, the sampling operation relies solely on the graph structure and can be performed in pre-processing. Finally, our empirical observations indicate that sampling low-order neighborhoods ($k=1,2$) is sufficient for achieving optimal transfer learning performance, which ensures computational efficiency in both sampling and pooling. We also provide a theoretical analysis to explain how subgraph information reduces the graph discrepancy.

\begin{theorem}
    For node $u \in \mathcal{V}^s$ in the source graph and $v \in \mathcal{V}^t$ in the target graph, considering the MEAN pooling function, the subgraph embeddings are $\mathbf{h}_u = \frac{\mathbf{z}_u + \sum_{i \in \mathcal{N}_s(u)} \mathbf{z}_i}{n + 1}$, $\mathbf{h}_v = \frac{\mathbf{z}_v + \sum_{j \in \mathcal{N}_s(v)} \mathbf{z}_j}{m + 1}$ where $n=|\mathcal{N}_s(u)|$, $m=|\mathcal{N}_s(v)|$. We have
    \begin{equation}
        \| \mathbf{h}_u - \mathbf{h}_v \| \le \| \mathbf{z}_u - \mathbf{z}_v \| - \Delta,
    \end{equation}
    where $\Delta = \frac{(n \| \mathbf{z}_u - \mathbf{z}_v \| - \frac{m - n}{m+1} \| \mathbf{z}_v \|)}{n+1}$ denotes the discrepancy margin.
\end{theorem}

\begin{corollary}
    If either of the following conditions is satisfied ($|\mathcal{N}_s(u)| \ge |\mathcal{N}_s(v)|$ or $|\mathcal{N}_s(u)|$ is sufficiently large), the inequality $\| \mathbf{h}_u - \mathbf{h}_v \| \le \| \mathbf{z}_u - \mathbf{z}_v \|$ strictly holds.
\end{corollary}

\begin{corollary}
    If the following condition is satisfied ($|\mathcal{N}_s(u)| < |\mathcal{N}_s(v)|$), the inequality $\| \mathbf{h}_u - \mathbf{h}_v \| \le \| \mathbf{z}_u - \mathbf{z}_v \|$ strictly holds when $\lambda \ge 2$, even in extreme case where $|\mathcal{N}_s(u)| \to 0$ and $|\mathcal{N}_s(v)| \to \infty$.
\end{corollary}

% \noindent\textit{Proof.} All proofs are presented in Appendix A.

\begin{remark}
    Based on the theoretical results, we can readily prove that the distance of $k$-th order central moment between two graphs can be reduced, i.e., $\| c_k(\mathbf{H}^s) - c_k(\mathbf{H}^t) \| \le \| c_k(\mathbf{Z}^s) - c_k(\mathbf{Z}^t) \|$. This implies that the SP layer indeed decreases the discrepancy (CMD value) between two graphs.
\end{remark}

\subsection{Subgraph Pooling++}

\begin{table*}[!t]
    \centering
    \resizebox{\linewidth}{!}{
        \begin{tabular}{c|l|cccc|cccc|c}
            \toprule
                                 &             & \multicolumn{4}{c|}{DBLP $\to$ ACM} & \multicolumn{4}{c|}{ACM $\to$ DBLP}                                                                                                                                                                          \\
            Backbone             & Model       & $q=0.1\%$                           & $q=0.5\%$                           & $q=1\%$                  & $q=10\%$                 & $q=0.1\%$                & $q=0.5\%$                & $q=1\%$                  & $q=10\%$                 & Rank \\ \midrule
            \multirow{7}{*}{GCN} & No Transfer & 48.44 ± 2.50                        & 62.70 ± 2.91                        & 68.63 ± 2.51             & 78.23 ± 0.41             & 39.12 ± 6.52             & 92.14 ± 1.77             & 95.61 ± 1.06             & 97.19 ± 0.18             & 4.8  \\ \cmidrule{2-11}
                                 & ERM         & 73.36 ± 0.88                        & 74.08 ± 0.67                        & 75.18 ± 0.54             & 76.19 ± 0.92             & 70.52 ± 0.91             & 80.88 ± 1.48             & 81.76 ± 0.74             & 83.07 ± 0.90             & 4.6  \\
                                 & Multi-task  & 70.10 ± 5.50                        & 70.96 ± 7.94                        & 74.35 ± 2.87             & 76.32 ± 2.79             & 74.51 ± 0.58             & 80.21 ± 0.97             & 80.24 ± 1.03             & 84.56 ± 1.04             & 5.3  \\
                                 & EERM        & 56.94 ± 6.49                        & 59.39 ± 6.33                        & 64.32 ± 6.93             & 67.96 ± 7.30             & 59.29 ± 6.23             & 70.10 ± 5.39             & 77.39 ± 3.05             & 90.03 ± 5.30             & 6.5  \\
                                 & GTrans      & 72.20 ± 0.19                        & 73.70 ± 1.93                        & 75.10 ± 0.11             & 77.53 ± 1.94             & 80.97 ± 1.84             & 88.84 ± 1.29             & 94.00 ± 3.09             & 95.19 ± 0.69             & 3.9  \\
                                 & GNN-SP      & \underline{74.51 ± 1.23}            & \underline{75.63 ± 1.61}            & \underline{75.64 ± 0.89} & \underline{79.18 ± 0.40} & \textbf{84.11 ± 2.00}    & \textbf{96.40 ± 1.65}    & \underline{96.41 ± 1.52} & \underline{97.54 ± 1.01} & 1.8  \\
                                 & GNN-SP++    & \textbf{74.68 ± 1.07}               & \textbf{76.41 ± 1.83}               & \textbf{77.06 ± 0.90}    & \textbf{79.20 ± 0.23}    & \underline{81.69 ± 5.96} & \underline{95.42 ± 2.74} & \textbf{96.66 ± 1.47}    & \textbf{98.20 ± 0.54}    & 1.3  \\ \midrule\midrule
            \multirow{7}{*}{GAT} & No Transfer & 48.11 ± 2.89                        & 62.52 ± 2.50                        & 68.50 ± 2.13             & 78.32 ± 0.32             & 40.30 ± 0.11             & 94.78 ± 2.24             & 96.68 ± 1.33             & 97.31 ± 0.28             & 4.9  \\\cmidrule{2-11}
                                 & ERM         & 68.48 ± 2.91                        & 72.60 ± 2.15                        & 72.67 ± 1.65             & 73.10 ± 2.02             & \underline{75.38 ± 2.32} & 85.76 ± 1.82             & 86.35 ± 1.42             & 87.99 ± 1.76             & 4.3  \\
                                 & Multi-task  & 67.72 ± 4.69                        & 69.37 ± 2.94                        & 70.72 ± 3.19             & 73.64 ± 3.80             & 71.34 ± 2.16             & 81.91 ± 2.73             & 81.74 ± 2.78             & 85.10 ± 2.99             & 6.0  \\
                                 & EERM        & 67.49 ± 2.89                        & 69.69 ± 1.85                        & 71.89 ± 2.48             & 74.48 ± 2.95             & 72.15 ± 2.91             & 79.80 ± 3.20             & 82.11 ± 1.34             & 89.48 ± 2.68             & 5.4  \\
                                 & GTrans      & 67.39 ± 2.09                        & 71.36 ± 0.49                        & 72.99 ± 1.34             & 75.36 ± 2.56             & 74.83 ± 1.92             & 85.03 ± 1.39             & 93.15 ± 1.54             & 95.59 ± 0.53             & 4.3  \\
                                 & GNN-SP      & \underline{70.85 ± 2.83}            & \underline{75.98 ± 1.03}            & \underline{76.56 ± 0.72} & \underline{78.56 ± 0.67} & \textbf{77.43 ± 6.47}    & \underline{95.44 ± 1.36} & \underline{96.55 ± 0.96} & \underline{97.63 ± 0.80} & 2.0  \\
                                 & GNN-SP++    & \textbf{71.88 ± 1.25}               & \textbf{76.27 ± 1.33}               & \textbf{77.14 ± 0.79}    & \textbf{79.02 ± 0.42}    & 75.14 ± 7.21             & \textbf{95.94 ± 1.37}    & \textbf{97.08 ± 1.07}    & \textbf{98.31 ± 0.20}    & 1.3  \\ \midrule\midrule
            \multirow{7}{*}{SGC} & No Transfer & 45.87 ± 5.79                        & 62.40 ± 2.77                        & 68.51 ± 2.41             & 78.49 ± 0.35             & 39.47 ± 3.88             & 92.37 ± 2.25             & 93.36 ± 1.10             & 96.13 ± 0.16             & 5.3  \\ \cmidrule{2-11}
                                 & ERM         & 73.44 ± 0.87                        & 74.37 ± 0.80                        & 74.63 ± 0.81             & 75.23 ± 1.04             & 70.07 ± 0.73             & 81.65 ± 1.61             & 81.43 ± 1.34             & 82.93 ± 0.89             & 4.8  \\
                                 & Multi-task  & 71.00 ± 0.62                        & 71.76 ± 1.11                        & 72.12 ± 2.39             & 74.71 ± 2.08             & 73.53 ± 1.16             & 79.35 ± 0.70             & 83.76 ± 1.06             & 84.27 ± 0.67             & 5.9  \\
                                 & EERM        & 72.45 ± 0.50                        & 72.95 ± 1.20                        & 74.55 ± 0.85             & 74.90 ± 0.58             & 74.35 ± 0.74             & 80.89 ± 0.58             & 82.12 ± 0.69             & 87.40 ± 0.45             & 4.8  \\
                                 & GTrans      & 71.72 ± 0.39                        & 72.02 ± 1.95                        & 73.97 ± 2.10             & 74.03 ± 0.50             & 80.29 ± 0.49             & 92.64 ± 0.61             & 93.68 ± 1.05             & 94.84 ± 1.54             & 4.4  \\
                                 & GNN-SP      & \underline{74.94 ± 1.24}            & \underline{75.67 ± 0.90}            & \underline{77.10 ± 1.12} & \textbf{79.35 ± 0.46}    & \underline{82.86 ± 1.06} & \textbf{96.07 ± 1.83}    & \underline{96.20 ± 1.53} & \underline{96.28 ± 1.48} & 1.8  \\
                                 & GNN-SP++    & \textbf{74.99 ± 0.57}               & \textbf{76.59 ± 1.72}               & \textbf{77.30 ± 1.30}    & \underline{79.15 ± 0.41} & \textbf{83.97 ± 1.93}    & \underline{95.75 ± 1.69} & \textbf{96.83 ± 1.70}    & \textbf{97.09 ± 1.67}    & 1.3  \\
            \bottomrule
        \end{tabular}}
    \scriptsize{$q$ denotes the ratio of training nodes in the target graph. For example, $q=10\%$ indicates 10 percent of nodes in the target graph are used for fine-tuning.}

    \caption{Node classification performance on \texttt{Citation} dataset. Rank indicates the average rank of all settings.}
    \label{tab:citation}
\end{table*}

The performance of Subgraph Pooling highly depends on the choice of subgraph sampling function. Employing an inappropriate sampling function can impair the distinguishability of the learned embeddings. For instance, the basic $k$-hop sampler might cause distinct nodes to share an identical subgraph, collapsing the embeddings into a single point. This can result in the potential over-smoothing~\cite{zhao2019pairnorm,keriven2022not,huang2020tackling}.

To address the limitation, we propose an advanced method Subgraph Pooling++ (SP++) that leverages Random Walk (RW)~\cite{huang2021broader} to sample subgraphs. We use the same hyper-parameter $k$ to define the maximum walk length, restricting the sampling process within $k$-hop subgraphs. The RW sampler is defined as
\begin{equation}
    \mathcal{N}_r(i) = \text{Sample}_{\text{RW}}(\mathcal{G}, i).
\end{equation}
The RW sampler mitigates the over-smoothing by imposing nodes to share different subgraphs. Inherently, $k$-hop sampler aims to cluster nodes with similar localized structural distributions. RW sampler further enhances the distinctiveness between structurally distant nodes, thereby creating more distinguishable clusters (Figure \ref{fig:rw sampling} \textbf{(Left)}). This improved distinguishability helps the classifier to capture meaningful information in prediction (Figure \ref{fig:rw sampling} \textbf{(Right)}).

Another approach to mitigate the risk of over-smoothing is to design advanced pooling functions. For example, we can employ attention mechanism~\cite{lee2019self} or hierarchical pooling~\cite{wu2022structural} to adaptively assign pooling weights $w_{ij}$ to nodes within a subgraph, thus preserving the uniqueness of subgraph embeddings. However, empirical evidence suggests that these pooling methods cannot offer significant advantages over basic MEAN pooling (Sec. \ref{sec:ablation}). Moreover, there is a concern regarding the efficiency of complicated pooling functions, as they could potentially increase computational and optimization efforts at each epoch. In contrast, the proposed RW sampling can be efficiently executed during pre-processing. To illustrate how RW sampling alleviates over-smoothing, we provide a concrete example below.

\begin{example}
    Considering two nodes $u, v \in \mathcal{V}^s$ in the source graph with $k$-hop sampler. Suppose $u, v$ share an identical $k$-hop subgraph yet different labels, i.e., $\mathcal{N}_s(u) = \mathcal{N}_s(u)$ and $y_i \neq y_j$, employing RW to sample neighborhoods $\mathcal{N}_r(u)$ and $\mathcal{N}_r(v)$, $\mathcal{N}_r(u) \neq \mathcal{N}_r(v)$,  can achieve lower empirical risk.
\end{example}

\begin{table*}[!t]
    \centering
    \resizebox{0.85\linewidth}{!}{
        \begin{tabular}{l|cc|cc|cc|c}
            \toprule
            Model       & Brazil $\to$ Europe               & USA $\to$ Europe                  & Brazil $\to$ USA                  & Europe $\to$ USA         & USA $\to$ Brazil         & Europe $\to$ Brazil      & Rank \\ \midrule
            % \# Nodes & \multicolumn{2}{c|}{399} & \multicolumn{2}{c|}{1190} & \multicolumn{2}{c|}{131} \\ 
            % \# Edges & \multicolumn{2}{c|}{12385} & \multicolumn{2}{c|}{28388} & \multicolumn{2}{c|}{2137} \\ \midrule\midrule
            No Transfer & \multicolumn{2}{c|}{48.63 ± 3.70} & \multicolumn{2}{c|}{59.18 ± 1.76} & \multicolumn{2}{c|}{52.36 ± 6.46} & 2.8                                                                                   \\ \midrule
            ERM         & 45.00 ± 2.95                      & 39.29 ± 3.96                      & 47.83 ± 1.92                      & 47.79 ± 4.66             & 39.53 ± 7.70             & 44.62 ± 4.24             & 6.8  \\
            Multi-task  & 48.55 ± 1.48                      & 47.61 ± 2.02                      & 48.73 ± 2.01                      & \underline{50.96 ± 2.12} & 52.17 ± 2.13             & 52.92 ± 6.07             & 4.3  \\
            EERM        & \underline{48.77 ± 2.85}          & 46.88 ± 4.70                      & \underline{48.91 ± 4.19}          & 48.36 ± 3.74             & 45.67 ± 3.68             & 46.65 ± 5.93             & 4.8  \\
            GTrans      & 48.50 ± 1.31                      & 47.49 ± 2.41                      & 48.84 ± 0.99                      & 48.88 ± 1.25             & 52.30 ± 1.50             & 53.00 ± 4.12             & 4.5  \\
            GNN-SP      & 48.76 ± 2.61                      & \textbf{51.30 ± 2.22}             & 46.06 ± 5.44                      & 49.85 ± 5.55             & \underline{55.47 ± 5.90} & \underline{54.72 ± 5.48} & 3.2  \\
            GNN-SP++    & \textbf{50.90 ± 3.93}             & \underline{50.40 ± 2.27}          & \textbf{51.06 ± 6.17}             & \textbf{53.87 ± 5.99}    & \textbf{57.08 ± 6.13}    & \textbf{55.23 ± 9.36}    & 1.5  \\ \bottomrule
        \end{tabular}}
    \caption{Node classification performance across \texttt{Airport} networks with GCN backbone. }
    \label{tab:airport}
\end{table*}

\paragraph{Illustration.} Let $\mathbf{h}_u = \sum_{i \in \mathcal{N}_s(u) \cup u} \mathbf{z}_i / (|\mathcal{N}_s(u)| + 1)$ and $\mathbf{h}_v = \sum_{j \in \mathcal{N}_s(v) \cup v} \mathbf{z}_j / (|\mathcal{N}_s(v)| + 1)$ as subgraph embeddings for node $u,v$ via MEAN pooling, where $\mathbf{h}_u = \mathbf{h}_v$. The empirical risk with classifier $g(\cdot)$ is given by:
\begin{equation}
    R_S = \frac{1}{2} \left( (g(\mathbf{h}_u) - y_u)^2 + (g(\mathbf{h}_v) - y_v)^2 \right).
\end{equation}
For simplicity, we use the mean square loss. $R_S$ is minimized when $g(\mathbf{h}_u) = y_u$ and $g(\mathbf{h}_v) = y_v$, but it is impossible to find a classifier $g(\cdot)$ to project a single vector into different labels. However, by applying RW to sample subgraphs with $\mathcal{N}_r(u)$ and $\mathcal{N}_r(v)$, we can obtain distinct subgraph embeddings  $\overline{\mathbf{h}}_u \neq \overline{\mathbf{h}}_v$. Then, it becomes feasible to find a classifier $g^*(\cdot)$ satisfying $g^*(\overline{\mathbf{h}}_u) = y_u$ and $g^*(\overline{\mathbf{h}}_v) = y_v$. In the extreme case, the subgraphs sampled by RW might be the same as $k$-hop, leading to $\overline{\mathbf{h}}_u = \overline{\mathbf{h}}_v$. To mitigate the issue, we can control the walk length and sampling frequency to maintain the distinctiveness of the sampled subgraphs. Therefore, utilizing RW sampler can lead to a lower empirical risk.

\section{Experiments}

\subsection{Experimental Setup}

\paragraph{Datasets.} We use \texttt{Citation}~\cite{wu2020unsupervised}, consisting of ACMv9 and DBLPv8; \texttt{Airport}~\cite{ribeiro2017struc2vec}, including Brazil, USA, and Europe; \texttt{Twitch}~\cite{rozemberczki2021multi} collected from six countries, including DE, EN, ES, FR, PT, RU; \texttt{Arxiv}~\cite{hu2020open} consisting papers with varying publish times; and dynamic financial network \texttt{Elliptic}~\cite{weber2019anti} that contains dozens of graph snapshots where each node is a Bitcoin transaction.

\paragraph{Baselines.} We include four GNN backbones: GCN~\cite{kipf2017semisupervised}, SAGE~\cite{hamilton2017inductive}, GAT~\cite{velivckovic2018graph}, and SGC~\cite{wu2019simplifying}. We compare our SP with No Transfer (directly training on target), Empirical Risk Minimization (transferring knowledge from source to target), Multi-task (jointly training on source and target), EERM~\cite{wu2022handling}, and recent SOTA method GTrans~\cite{jin2023empowering}. We also compare various domain adaptation methods, including DANN~\cite{ganin2016domain}, CDAN~\cite{long2018conditional}, UDAGCN~\cite{wu2020unsupervised}, MIXUP~\cite{wang2021mixup}, EGI~\cite{zhu2021transfer}, SR-GNN~\cite{zhu2021shift}, GRADE~\cite{wu2023non}, SSReg~\cite{you2023graph}, and StruRW~\cite{liu2023structural}.

\paragraph{Settings.} We pre-train the model on the source with 60 percent of labeled nodes and adapt the model to the target. The adaptation involves three settings: (1) directly applying the pre-trained model without any fine-tuning (Without FT); (2) fine-tuning the last layer (classifier) of the model (FT Last Layer); (3) fine-tuning all parameters (Fully FT). We take split 10/10/80 to form train/valid/test sets on the target graph. 
% Specific hyper-parameters are presented in the Appendix C.

\begin{figure}[!t]
    \centering
    \includegraphics[width=\linewidth]{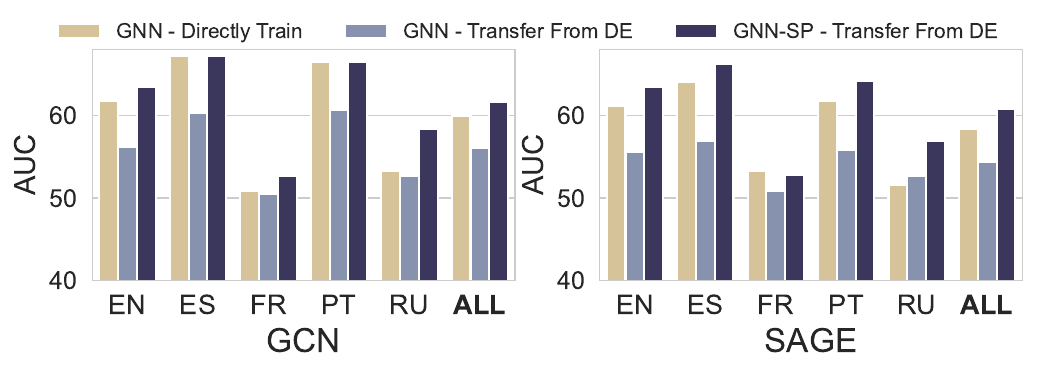}
    \caption{Node classification performance on \texttt{Twitch}. }
    \label{fig:twitch}
\end{figure}

\begin{figure}[!t]
    \centering
    \includegraphics[width=\linewidth]{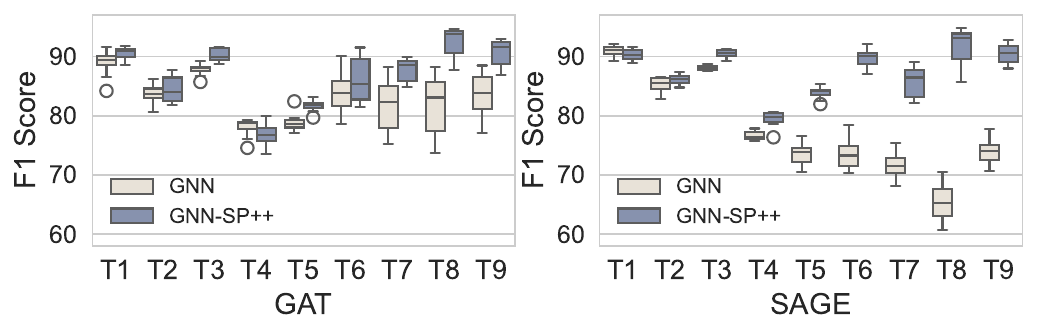}
    \caption{Node classification performance on \texttt{Elliptic}.}
    \label{fig:elliptic}
\end{figure}

\subsection{Node Classification}

\paragraph{One-to-One Transfer.} The transfer learning results on \texttt{Citation} over three GNN backbones are presented in Table \ref{tab:citation}. Considering the limited number of parameters, we apply the FT Last Layer setting. Our GNN-SP outperforms baselines across all settings and is even better than the model trained from scratch (No Transfer), demonstrating the capability of overcoming negative transfer. 
%%%%%%%%%%%%%%% Present in extended version %%%%%%%%%%%%%%%
Note that GNN-SP performs better than the advanced GNN-SP++ with extremely limited labels. It is may because (1) the dataset is small and sparse ($\sim$5k to $\sim$7k nodes and $\sim$20k to $\sim$30k edges) where RW sampler fails to model the real localized structures; and (2) the label sparsity exacerbates overfitting.
%%%%%%%%%%%%%%% Present in extended version %%%%%%%%%%%%%%%
The performance on \texttt{Airport} is presented in Table \ref{tab:airport}. Our method surpasses all baselines and achieves an average Rank of 1.5, which is notably higher than the No Transfer. Note that transferring knowledge to \texttt{USA} results in negative transfer for all baselines, likely due to its significantly larger size that potentially contains more patterns compared to the other two graphs.
%%%%%%%%%%%%%%% Present in extended version %%%%%%%%%%%%%%%
Specifically, \texttt{USA} contains 1,190 nodes and 28,388 edges, whereas \texttt{Europe} and \texttt{Brazil} contain only 399 nodes \& 12,385 edges, and 131 nodes \& 2,137 edges, respectively. The smaller graphs provide limited patterns and may introduce unexpected biases during pre-training. Additionally, we observe that GNN-SP++ performs better than GNN-SP, likely because these graphs are densely connected, leading to the issue that various nodes share identical $k$-hop subgraphs.
%%%%%%%%%%%%%%% Present in extended version %%%%%%%%%%%%%%%

\begin{table}[!t]
    \resizebox{\linewidth}{!}{
        \begin{tabular}{l|cccc|c}
            \toprule
            Model       & Time 1                            & Time 2                   & Time 3                   & Time 4                   & Rank \\ \midrule
            No Transfer & \multicolumn{4}{c|}{69.60 ± 0.31} & 2.8                                                                                   \\ \midrule
            ERM         & 65.73 ± 0.57                      & 66.18 ± 0.48             & 68.67 ± 0.32             & 70.33 ± 0.29             & 4.5  \\
            Multi-task  & 50.32 ± 2.17                      & 52.77 ± 2.82             & 60.02 ± 0.99             & 67.62 ± 0.75             & 6.8  \\
            EERM        & 55.25 ± 2.03                      & 57.47 ± 0.59             & 63.25 ± 0.54             & 65.26 ± 0.63             & 6.3  \\
            GTrans      & 65.95 ± 0.12                      & 66.64 ± 0.51             & \underline{69.51 ± 0.39} & \underline{71.54 ± 0.30} & 3.3  \\
            GCN-SP      & \underline{67.76 ± 0.23}          & \underline{68.36 ± 0.33} & 69.03 ± 0.63             & 69.75 ± 0.56             & 3.5  \\
            GCN-SP++    & \textbf{71.43 ± 0.52}             & \textbf{72.75 ± 1.24}    & \textbf{74.04 ± 0.83}    & \textbf{75.17 ± 0.21}    & 1.0  \\ \bottomrule
        \end{tabular}}
    \caption{Node classification on \texttt{Arxiv} with GCN backbone.}
    \label{tab:arxiv}
\end{table}

\begin{table*}[!ht]
    \centering
    \resizebox{0.75\linewidth}{!}{
        \begin{tabular}{l|ccccc}
            \toprule
                                   & ACM $\to$ DBLP           & DBLP $\to$ ACM           & Twitch-All               & Arxiv-T1                 & Arxiv-T3                 \\ \midrule
            GCN + Fully FT         & 97.75 ± 0.16             & \underline{80.03 ± 0.22} & 60.59 ± 1.13             & 69.29 ± 0.16             & 69.70 ± 0.39             \\
            GCN-SP + FT Last Layer & \underline{98.20 ± 0.54} & 79.20 ± 0.73             & \underline{61.66 ± 0.92} & \underline{71.43 ± 0.52} & \underline{74.04 ± 0.83} \\
            GCN-SP + Fully FT      & \textbf{98.66 ± 0.29}    & \textbf{80.82 ± 0.59}    & \textbf{61.77 ± 0.98}    & \textbf{73.12 ± 0.93}    & \textbf{75.01 ± 0.62}    \\ \bottomrule
        \end{tabular}}
    \caption{Comparison between fine-tuning the model classifier (FT Last Layer) and fine-tuning the whole model (Fully FT).}
    \label{tab:fully FT}
\end{table*}

\paragraph{One-to-Multi Transfer.} We use \texttt{Twitch} as the benchmark, which consists of six graphs with different sizes and data distributions. We pre-train the model on \texttt{DE} and fine-tune on other graphs (\texttt{EN}, \texttt{ES}, \texttt{FR}, \texttt{PT}, \texttt{RU}). We employ ROC-AUC as metric and adopt 2-layer GCN and SAGE as backbones. Figure \ref{fig:twitch} shows GNN-SP outperforms standard GNN with up to 8\% improvements on ROC-AUC under FT Last Layer setting and achieves better performance than the model directly trained on the target over 10 out of 12 settings. The results validate the generalizability of GNN-SP to multiple graphs.

\paragraph{Transfer with Dynamic Shift.} In this scenario, we evaluate the model's capability to handle temporal distribution shifts. We first adopt a dynamic financial network \texttt{Elliptic} with splitting 5/5/33 snapshots for train/valid/test. Figure \ref{fig:elliptic} presents the results where the test snapshots are grouped into 9 folds. The performance gain of our GNN-SP is up to 10\% and 24\% over GAT and SAGE, respectively. Additionally, we use \texttt{Arxiv} as another temporal dataset, where nodes represent papers published from 2005 to 2020 and edges indicate citations. Based on the publication time, we collect five sub-graphs, represented as \texttt{Time 1} (2005 - 2007), \texttt{Time 2} (2008 - 2010), \texttt{Time 3} (2011 - 2014), \texttt{Time 4} (2015 - 2017), and \texttt{Time 5} (2018 - 2020). We use the first four graphs as sources and the last one as the target. The results are presented in Table \ref{tab:arxiv} where our GNN-SP++ achieves significant improvements over all baselines. 
%%%%%%%%%%%%%%% Present in extended version %%%%%%%%%%%%%%%
We note that the temporal distribution shift is marginal when the source and target are temporally proximate, resulting in improved transfer learning performance. Additionally, the performance of GNN-SP++ is considerably superior to GNN-SP, further validating the efficacy of the RW sampler.
%%%%%%%%%%%%%%% Present in extended version %%%%%%%%%%%%%%%

\subsection{Ablation Study}
\label{sec:ablation}

\paragraph{Transfer without Fine-tuning.} Following~\cite{liu2023structural}, we take a further step by directly employing the pre-trained model on target graph without fine-tuning. Table \ref{tab:without FT} presents the transfer learning performance on \texttt{Citation} and \texttt{Arxiv}. We also adopt another domain adaptation setting (Degree) from~\cite{gui2022good}. It is obvious that our proposed SP layer significantly improves the transfer learning performance by enhancing the quality of the encoder.

\begin{table}[!t]
    \resizebox{\linewidth}{!}{
        \begin{tabular}{l | cc | cc}
            \toprule
                       & \multicolumn{2}{c}{ACM \& DBLP} & \multicolumn{2}{|c}{Arxiv}                                                       \\
            Model      & A $\to$ D                       & D $\to$ A                  & Time 1                   & Degree                   \\ \midrule
            GCN$^*$    & 59.02 ± 1.04                    & 59.20 ± 0.70               & 28.08 ± 0.24             & 57.41 ± 0.14             \\
            GAT$^*$    & 61.67 ± 3.54                    & 62.18 ± 7.04               & 32.32 ± 1.10             & 58.10 ± 0.15             \\ \midrule
            DANN       & 59.02 ± 7.79                    & 65.77 ± 0.46               & 24.33 ± 1.19             & 56.13 ± 0.18             \\
            CDAN       & 60.56 ± 4.38                    & 64.35 ± 0.83               & 25.85 ± 1.15             & 56.43 ± 0.45             \\
            UDAGCN     & 59.62 ± 2.86                    & 64.74 ± 2.51               & 25.64 ± 3.04             & 55.77 ± 0.83             \\
            EERM       & 40.88 ± 5.10                    & 51.71 ± 5.07               & -                        & -                        \\
            MIXUP      & 49.93 ± 0.89                    & 63.36 ± 0.66               & 28.04 ± 0.18             & \underline{59.22 ± 0.22} \\
            EGI$^*$    & 49.03 ± 1.50                    & 64.40 ± 1.03               & 25.59 ± 0.25             & 56.93 ± 0.23             \\
            SR-GNN$^*$ & 62.49 ± 1.96                    & 63.32 ± 1.49               & 25.44 ± 0.30             & 56.98 ± 0.12             \\
            GRADE$^*$  & 67.29 ± 2.04                    & 64.13 ± 3.12               & 25.69 ± 0.12             & 57.49 ± 0.39             \\
            SSReg$^*$  & 69.04 ± 2.95                    & \underline{65.93 ± 1.05}   & 27.93 ± 0.29             & 56.67 ± 0.33             \\
            StruRW     & \underline{70.19 ± 2.10}        & 65.07 ± 1.98               & \underline{28.46 ± 0.18} & 57.45 ± 0.15             \\ \midrule
            GCN-SP++   & \textbf{75.88 ± 5.57}           & \textbf{71.32 ± 1.33}      & \textbf{40.41 ± 1.07}    & 64.35 ± 0.41             \\
            GAT-SP++   & 73.78 ± 8.13                    & 67.05 ± 4.97               & 36.38 ± 2.50             & \textbf{65.53 ± 0.60}    \\ \bottomrule
        \end{tabular}
    }
    \scriptsize{The reported results are from~\cite{liu2023structural}. $^*$ indicates the results of our implementations based on the official code.}

    \caption{Node classification results without fine-tuning.}
    \label{tab:without FT}
\end{table}

\paragraph{Transfer with Fully Fine-tuning.} We present transfer learning performance by fine-tuning the whole model in Table \ref{tab:fully FT}. We observe even if GNN-SP only fine-tunes the last layer, it still outperforms standard GNN with full fine-tuning. If our GNN-SP is fully fine-tuned, the model performance can be further improved, especially on large-scaled \texttt{Arxiv}.

\begin{figure}[!t]
    \centering
    \includegraphics[width=\linewidth]{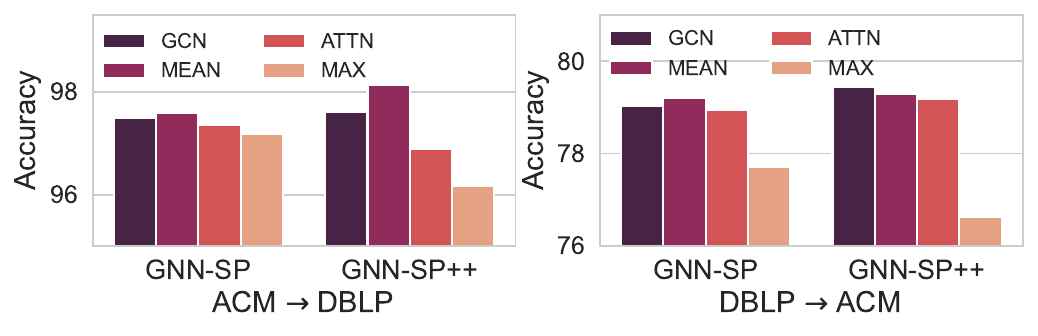}
    \caption{Ablation on \texttt{Citation} with different pooling functions.}
    \label{fig:ablation pooling}
\end{figure}

\paragraph{Pooling Methods.} We evaluate model performance with different pooling functions, including MEAN, ATTN, MAX, and GCN. Figure \ref{fig:ablation pooling} presents the experimental results over \texttt{Citation} network with GCN backbone. We observe the basic MEAN pooling outperforms complicated GCN and ATTN that adaptively determine the pooling weights $w_{ij}$. It is might because the complicated methods introduce extra inductive bias with unexpected noises.

\section{Related Works}

Existing studies are typically categorized based on the availability of the target graph during the pre-training phase.

\paragraph{Pre-training with Target Graph.} Researchers developed methods to explicitly align source and target graphs during pre-training, ensuring $P_S(\mathcal{Z}) = P_T(\mathcal{Z})$. For example,~\cite{zhang2019dane,dai2022graph} adopt adversarial learning~\cite{ganin2016domain} to train domain-invariant encoder, and following UDAGCN~\cite{wu2020unsupervised} incorporates attention to further enhance expressiveness. Alternatively, one can employ regularizers~\cite{zhu2021shift,zhu2023explaining,shi2023improving}, such as MMD and CMD, to constrain the discrepancy between the source and target. To facilitate this process, new graph-specific discrepancy metrics are proposed, including tree mover distance~\cite{chuang2022tree}, subtree discrepancy~\cite{wu2023non}, and spectral regularizer~\cite{you2023graph}. Additionally,~\cite{liu2023structural} emphasizes the most relevant instances in the source to better match distributions of the source and target. While these methods effectively reduce the distribution shift between source and target, the target graph may be unavailable during pre-training in many real-world scenarios.

\paragraph{Pre-training without Target Graph.} Existing works aim to train GNNs transferable to unseen graphs. For example, EERM~\cite{wu2022handling} and following works~\cite{chen2022learning,wu2022discovering,wu2023energybased,yu2023mind} utilize causal learning to develop environment-invariant encoder. GTrans~\cite{jin2023empowering} transforms the target graph at test-time to align the source and target. Moreover, various studies employ augmentation to enhance the robustness of encoder against permutations~\cite{verma2021graphmix,wang2021mixup,liu2022local,han2022g,liu2023good,guo2023data} or apply disentangle learning to extract domain-invariant semantics~\cite{ma2019disentangled,liu2020independence}. To understand the transferability of GNNs,~\cite{ruiz2020graphon,levie2021transferability,cao2023pre} interpret graphs as the combination of graphons, while~\cite{han2021adaptive,zhu2021transfer,sun2023all,qiu2020gcc} adopt self-supervised learning to identify transferable structures. Despite these efforts to enhance transferability, they cannot well address the negative transfer issue. To this end, we systematically analyze why the negative transfer happens and provide insights to solve this issue.
% More recently, OFA~\cite{liu2023one} leverages LLMs to align graphs from different domains to a unified space, enabling cross-domain and cross-task transferring, which is complementary to our work.

\section{Conclusion}

In this paper, we analyze the negative transfer in GNNs and introduce Subgraph Pooling, a simple yet effective method, to mitigate the issue. Our method transfers the subgraph-level knowledge to reduce the discrepancy between the source and target graphs, and is applicable for any GNN backbone without introducing extra parameters. We provide a theoretical analysis to demonstrate how the model works and conduct extensive experiments to evaluate its superiority under various transfer learning settings.

\section*{Acknowledgements}

This work was partially supported by the NSF under grants IIS-2321504, IIS-2334193, IIS-2340346, IIS-2203262, IIS-2217239, CNS-2203261, and CMMI-2146076. Any opinions, findings, and conclusions or recommendations expressed in this material are those of the authors and do not necessarily reflect the views of the sponsors.

%% The file named.bst is a bibliography style file for BibTeX 0.99c
\bibliographystyle{named}
\bibliography{ijcai24}

\appendix
\onecolumn

\section{Proof}

\noindent\textbf{Theorem 1.} \textit{For node $u \in \mathcal{V}^s$ in the source graph and $v \in \mathcal{V}^t$ in the target graph, considering the MEAN pooling function, the subgraph embeddings are $\mathbf{h}_u = \frac{\mathbf{z}_u + \sum_{i \in \mathcal{N}_s(u)} \mathbf{z}_i}{n + 1}$, $\mathbf{h}_v = \frac{\mathbf{z}_v + \sum_{j \in \mathcal{N}_s(v)} \mathbf{z}_j}{m + 1}$ where $n=|\mathcal{N}_s(u)|$, $m=|\mathcal{N}_s(v)|$. We have
    \begin{equation}
        \| \mathbf{h}_u - \mathbf{h}_v \| \le \| \mathbf{z}_u - \mathbf{z}_v \| - \Delta,
    \end{equation}
    where $\Delta = \frac{(n \| \mathbf{z}_u - \mathbf{z}_v \| - \frac{m - n}{m+1} \| \mathbf{z}_v \|)}{n+1}$ denotes the discrepancy margin.}

% \noindent\textit{Proof.} 
\begin{proof}
To demonstrate this theorem, we analyze the distance between the subgraph embeddings $\| \mathbf{h}_u - \mathbf{h}_v \|$ in terms of the node embeddings:
\begin{align}
    % & \quad\quad \| \mathbf{h}_u - \mathbf{h}_v \| \\
    \left\| \frac{\mathbf{z}_u + \sum_{i \in \mathcal{N}_s(u)} \mathbf{z}_i}{n + 1} - \frac{\mathbf{z}_v + \sum_{j \in \mathcal{N}_s(v)} \mathbf{z}_j}{m + 1} \right\| & = \left\| \frac{(m+1)(\mathbf{z}_u + \sum_{i} \mathbf{z}_i) - (n+1) (\mathbf{z}_v + \sum_{j} \mathbf{z}_j)}{(m+1)(n+1)} \right\|                                                                                      \\
                                                                                                                                                                       & = \left\| \frac{(m+1)\mathbf{z}_u - (n+1) \mathbf{z}_v + (m+1) \sum_{i} \mathbf{z}_i - (n+1) \sum_{j} \mathbf{z}_j)}{(m+1)(n+1)} \right\|                                                                             \\
                                                                                                                                                                       & \le \underbrace{\left\| \frac{(m+1)\mathbf{z}_u - (n+1) \mathbf{z}_v}{(m+1)(n+1)} \right\|}_{(a)} + \underbrace{\left\| \frac{(m+1) \sum_{i} \mathbf{z}_i - (n+1) \sum_{j} \mathbf{z}_j)}{(m+1)(n+1)} \right\|}_{(b)}
\end{align}
This inequality results from the triangle inequality. Simplifying these terms separately, we have the term (a) as:
\begin{align}
    \left\| \frac{(m+1)\mathbf{z}_u - (n+1) \mathbf{z}_v}{(m+1)(n+1)} \right\| & = \frac{\left\| (m+1) (\mathbf{z}_u - \mathbf{z}_v) + (m-n) \mathbf{z}_v \right\|}{(m+1)(n+1)}                                                                                                \\
                                                                               & \le \frac{\left\| \mathbf{z}_u - \mathbf{z}_v \right\|}{n+1} + \frac{(m-n) \left\| \mathbf{z}_v \right\|}{(m+1)(n+1)}                                                                         \\
                                                                               & = \left\| \mathbf{z}_u - \mathbf{z}_v \right\| - \left\| \mathbf{z}_u - \mathbf{z}_v \right\| + \frac{\left\| \mathbf{z}_u - \mathbf{z}_v \right\| + \frac{m-n}{m+1} \| \mathbf{z}_v \|}{n+1} \\
                                                                               & = \left\| \mathbf{z}_u - \mathbf{z}_v \right\| - \frac{(n \| \mathbf{z}_u - \mathbf{z}_v \| - \frac{m - n}{m+1} \| \mathbf{z}_v \|)}{n+1}                                                     \\
                                                                               & = \left\| \mathbf{z}_u - \mathbf{z}_v \right\| - \Delta
\end{align}
where $\Delta = \frac{(n \| \mathbf{z}_u - \mathbf{z}_v \| - \frac{m - n}{m+1} \| \mathbf{z}_v \|)}{n+1}$ indicates the discrepancy gap handled by our proposed Subgraph Pooling. Additionally, the term (b) can be simplified as:
\begin{align}
    \left\| \frac{(m+1) \sum_{i \in \mathcal{N}_s(u)} \mathbf{z}_i - (n+1) \sum_{j \in \mathcal{N}_s(v)} \mathbf{z}_j)}{(m+1)(n+1)} \right\| & = \left\| \frac{\sum_{i \in \mathcal{N}_s(u)} \mathbf{z}_i}{n+1} - \frac{\sum_{j \in \mathcal{N}_s(v)} \mathbf{z}_j}{m+1} \right\| \le \epsilon
\end{align}
For this term, under Assumption 1, it is sufficiently small to be disregarded. Combining these analyses, we derive $\| \mathbf{h}_u - \mathbf{h}_v \| \le \| \mathbf{z}_u - \mathbf{z}_v \| - \Delta$. While $\Delta$ may not always be positive, it is essential for understanding the impact of SP. To further substantiate this, we present two corollaries.
\end{proof}

\noindent\textbf{Corollary 1.} \textit{If either of the following conditions is satisfied ($|\mathcal{N}_s(u)| \ge |\mathcal{N}_s(v)|$ or $|\mathcal{N}_s(u)|$ is sufficiently large), the inequality $\| \mathbf{h}_u - \mathbf{h}_v \| \le \| \mathbf{z}_u - \mathbf{z}_v \|$ strictly holds.}

% \noindent\textit{Proof.} 
\begin{proof}
To establish that $\Delta \ge 0$, we need to demonstrate that:
\begin{align}
    n \| \mathbf{z}_u - \mathbf{z}_v \|                          & \ge \frac{m - n}{m+1} \| \mathbf{z}_v \| \\
    \frac{\| \mathbf{z}_u - \mathbf{z}_v \|}{\| \mathbf{z}_v \|} & \ge \frac{(m-n)}{n(m+1)},
\end{align}
where $n = |\mathcal{N}_s(u)|$ and $m = |\mathcal{N}_s(v)|$.

We consider the following two cases:
\begin{enumerate}
    \item In cases where the source graph is richer than the target, namely, $|\mathcal{N}_s(u)| \ge |\mathcal{N}_s(v)|$, we have $\frac{(m-n)}{n(m+1)} \le 0$. Under these conditions, $\frac{\| \mathbf{z}_u - \mathbf{z}_v \|}{\| \mathbf{z}_v \|} \ge \frac{(m-n)}{n(m+1)}$ is strictly valid, given that $\| \mathbf{z}_u - \mathbf{z}_v \| \ge 0$ and $\| \mathbf{z}_v \| \ge 0$.

    \item When both source and target graphs are substantially rich, $\frac{(m-n)}{n(m+1)} \le 0$ remains strictly valid. In the extreme case where $n \to \infty$ and $m \to \infty$, it follows that $\lim_{n \to \infty, m \to \infty} \frac{(m-n)}{n(m+1)} = 0$. This suggests that when the source graph is adequately rich, essential patterns can be transferred effectively, regardless of the scale of the target.
\end{enumerate}

\end{proof}

\noindent\textbf{Corollary 2.} \textit{If the following condition is satisfied ($|\mathcal{N}_s(u)| < |\mathcal{N}_s(v)|$), the inequality $\| \mathbf{h}_u - \mathbf{h}_v \| \le \| \mathbf{z}_u - \mathbf{z}_v \|$ strictly holds when $\lambda \ge 2$, even in extreme case where $|\mathcal{N}_s(u)| \to 0$ and $|\mathcal{N}_s(v)| \to \infty$.}

% \noindent\textit{Proof.} 
\begin{proof}
Assuming the target graph is richer than the source, indicated by $|\mathcal{N}_s(v)| \ge |\mathcal{N}_s(u)|$, it is necessary to validate that $\| \mathbf{z}_u - \mathbf{z}_v \| \ge \| \mathbf{z}_v \|$ to maintain the inequality. This validation becomes crucial in the extreme case where $n \to 0$ and $m \to \infty$, leading to $\lim_{n \to 0, m \to \infty} \frac{(m-n)}{n(m+1)} = 1$. We must demonstrate that $\frac{\| \mathbf{z}_u - \mathbf{z}_v \|}{\| \mathbf{z}_v \|} \ge \frac{(m-n)}{n(m+1)}$ is valid under these circumstances, implying $\frac{\| \mathbf{z}_u - \mathbf{z}_v \|}{\| \mathbf{z}_v \|} \ge 1$. Therefore, when the source is sparse, it may lack sufficient information for effective transfer.
\begin{align}
     & \| \mathbf{z}_u - \mathbf{z}_v \| \ge \| \mathbf{z}_v \|                                                                              \\
     & \Rightarrow \| \mathbf{z}_u - \mathbf{z}_v \|^2 \ge \| \mathbf{z}_v \|^2                                                              \\
     & \Rightarrow\| \mathbf{z}_u \|^2 + \| \mathbf{z}_v \|^2 - 2 \| \mathbf{z}_u \| \| \mathbf{z}_v \| \cos \theta \ge \| \mathbf{z}_v \|^2 \\
     & \Rightarrow \| \mathbf{z}_u \|^2 \ge 2 \| \mathbf{z}_u \| \| \mathbf{z}_v \| \cos \theta                                              \\
    % & \Rightarrow \| \mathbf{z}_u \|^2 \ge 2 \mathbf{z}_u^T \mathbf{z}_v \\
    % & \Rightarrow \mathbf{z}_u^T \mathbf{z}_u \ge 2 \mathbf{z}_u^T \mathbf{z}_v \\
    % & \Rightarrow \frac{\| \mathbf{z}_u \|^2}{\| \mathbf{z}_u \| \| \mathbf{z}_v \| \cos \theta} \ge 2 \\ 
     & \Rightarrow \frac{\mathbf{z}_u^T \mathbf{z}_u}{\mathbf{z}_u^T \mathbf{z}_v} \ge 2
\end{align}
where $\cos \theta = \mathbf{z}_x^T \mathbf{z}_y / \| \mathbf{z}_x \| \| \mathbf{z}_y \|$. According to Definition 2, this inequality is satisfied when $\lambda \ge 2$.
\end{proof}

\section{Dataset Details}

\begin{table*}[!h]
    \centering
    \resizebox{\linewidth}{!}{
        \begin{tabular}{ccccccccc}
            \toprule
            Dataset                            & Setting              & \# Nodes & \# Edges  & \# Classes & \# Density & Avg Degree & Max Degree & Metric                    \\ \midrule
            \multirow{2}{*}{\texttt{Citation}} & ACMv9                & 7,410    & 29,456    & 6          & 0.00054    & 3.98       & 108        & \multirow{2}{*}{Accuracy} \\
                                               & DBLPv8               & 5,578    & 20,158    & 6          & 0.00065    & 3.61       & 254        &                           \\ \midrule
            \multirow{3}{*}{\texttt{Airport}}  & USA                  & 1,190    & 28,388    & 4          & 0.02006    & 23.86      & 239        & \multirow{3}{*}{Accuracy} \\
                                               & Europe               & 399      & 12,385    & 4          & 0.07799    & 31.04      & 203        &                           \\
                                               & Brazil               & 131      & 2,137     & 4          & 0.12548    & 16.31      & 80         &                           \\ \midrule
            \multirow{6}{*}{\texttt{Twitch}}   & DE                   & 9,498    & 315,774   & 2          & 0.00350    & 33.25      & 4,260      & \multirow{6}{*}{ROC-AUC}  \\
                                               & EN                   & 7,126    & 77,774    & 2          & 0.00153    & 10.91      & 721        &                           \\
                                               & ES                   & 4,648    & 123,412   & 2          & 0.00571    & 26.55      & 1,023      &                           \\
                                               & FR                   & 6,551    & 231,883   & 2          & 0.00540    & 35.40      & 2,041      &                           \\
                                               & PT                   & 1,912    & 64,510    & 2          & 0.01766    & 33.74      & 768        &                           \\
                                               & RU                   & 4,385    & 78,993    & 2          & 0.00411    & 18.01      & 1,230      &                           \\ \midrule
            \multirow{6}{*}{\texttt{Arxiv}}    & Time 1 [2005 - 2007] & 4,980    & 10,086    & 40         & 0.00041    & 2.03       & 14         & \multirow{6}{*}{Accuracy} \\
                                               & Time 2 [2008 - 2010] & 12,974   & 32,215    & 40         & 0.00019    & 2.48       & 43         &                           \\
                                               & Time 3 [2011 - 2014] & 41,125   & 140,526   & 40         & 0.00008    & 3.42       & 76         &                           \\
                                               & Time 4 [2015 - 2017] & 90,941   & 462,438   & 40         & 0.00006    & 5.09       & 222        &                           \\
                                               & Time 5 [2018 - 2020] & 169,343  & 1,335,586 & 40         & 0.00005    & 7.89       & 437        &                           \\
                                               & Degree               & 169,343  & 2,484,941 & 40         & 0.00009    & 14.67      & 13,162     &                           \\ \bottomrule
        \end{tabular}
    }
    \caption{The statistics of \texttt{Citation}, \texttt{Airport}, \texttt{Twitch}, and \texttt{Arxiv} networks.}
    \label{tab:statistics}
\end{table*}

\begin{figure*}[!h]
    \centering
    \subfloat{\includegraphics[width=\linewidth]{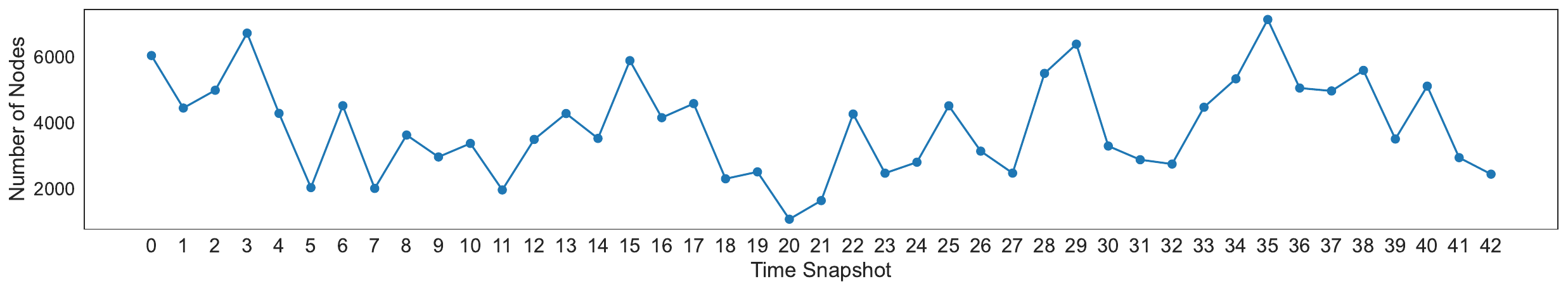}} \\
    \subfloat{\includegraphics[width=\linewidth]{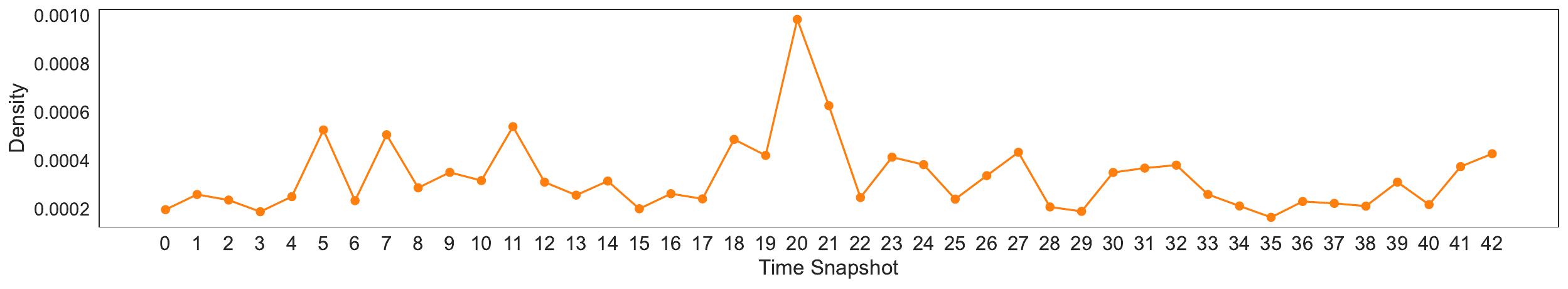}} \\
    \subfloat{\includegraphics[width=\linewidth]{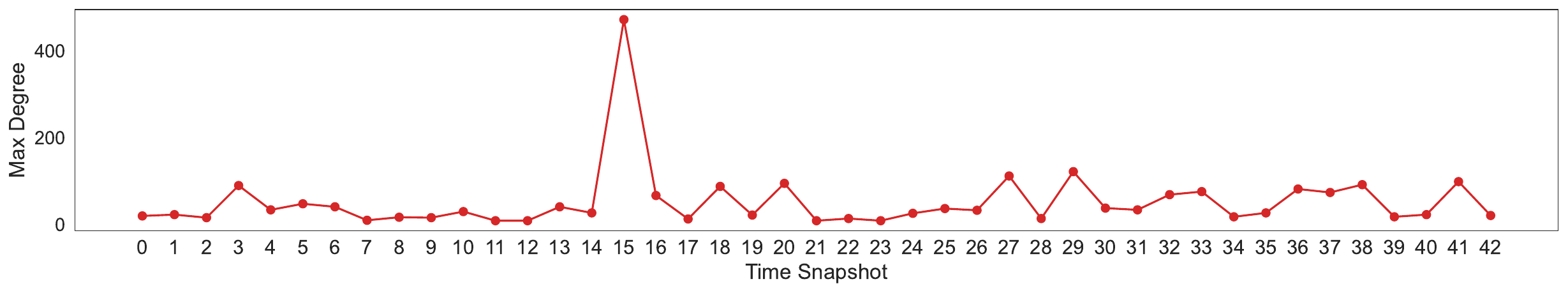}}
    \caption{Statistics of \texttt{Elliptic} dataset.}
    \label{fig:statistics elliptic}
\end{figure*}

In this section, we detail the datasets utilized in our experiments, with their statistics summarized in Table \ref{tab:statistics}.

\begin{itemize}
    \item \texttt{Citation}. The Citation networks consist of the ACM and DBLP datasets, each sourced from distinct academic databases. In these networks, nodes correspond to academic papers, while edges represent citations between them. Both ACM and DBLP demonstrate comparable sizes and densities. The datasets are categorized into six research fields, namely Database, Data Mining, Artificial Intelligence, Computer Vision, Information Security, and High Performance Computing. Despite the scale of DBLP is relatively smaller compared to ACM, it has a higher maximum degree of 254, as opposed to 108 of ACM. This difference is attributed to the variation in structural distribution between the two datasets.
    \item \texttt{Airport}. The Airport networks include three distinct datasets from various regions, where nodes symbolize airports, and edges represent flight connections. The labels in these datasets indicate the airport activity level, quantified by the number of flights or passengers. Compared to other datasets, the density of the Airport is much higher where the number of edges is much higher than the number of nodes. It is noteworthy that the USA dataset might encapsulate more intricate patterns due to its larger size compared to Europe and Brazil.
    \item \texttt{Twitch}. Twitch dataset contains six networks from different regions, including DE, EN, ES, FR, PT, and RU. In these networks, nodes correspond to users on the streaming platform, and edges reflect friendships. The primary task is to predict mature content within the network. The model is initially pre-trained on the comparatively larger DE network and subsequently transferred to other networks, each possessing distinct distributions on network scales and densities. These networks are leveraged to test model expressiveness on various target graphs.
    \item \texttt{Arxiv}. Arxiv is another citation network where nodes represent papers published from 2005 to 2020, and edges correspond to citations between these papers. The network is released by OGB. We divided the original dataset into five distinct sub-datasets, each segmented according to the time of paper publication, shown in Table \ref{tab:statistics}. We transfer knowledge from the previous four time periods to the final one, each characterized by differing temporal distributions. Generally, temporally close networks share relatively similar distributions.
    \item \texttt{Elliptic}. Elliptic consists of 43 snapshots, where the dataset distributions of each one are illustrated in Figure \ref{fig:statistics elliptic}. Specifically, we present the number of nodes, density, and the max degree of each snapshot. We can observe a clear temporal distribution shift across these graphs. We use a 5/5/33 split for train/valid/test sets to evaluate the model robustness on long-range temporal distribution shift.
\end{itemize}

\section{Hyper-parameters}

To prevent the impact of randomness, we run each model 10 times and report the mean and standard deviation. For all baseline models, we standardized the hidden dimension to 64, the learning rate to 0.001 (1e-3), and the number of backbone layers to 2. In the case of attention-based methods, we configured the number of attention heads to 4. For additional configurations, we adhered to the specifications outlined in their respective original papers. The comprehensive hyper-parameter setting for our proposed Subgraph Pooling method is detailed in Table \ref{tab:hyperparameter}.

\begin{table*}[!h]
    \centering
    \resizebox{\linewidth}{!}{
        \begin{tabular}{lcccccc}
            \toprule
                                    & \texttt{Citation}                               & \texttt{Airport} & \texttt{Twitch} & \texttt{Arxiv} & \texttt{Elliptic} & \texttt{Facebook} \\ \midrule
            Hidden Dimension        & \multicolumn{6}{c}{64 for all datasets}                                                                                                       \\
            Activation              & \multicolumn{6}{c}{PReLU used for all datasets}                                                                                               \\
            \# Encoder Layers       & 2                                               & 2                & 2               & 2              & 2                 & 5                 \\
            Normalization           & \multicolumn{6}{c}{-}                                                                                                                         \\
            Pre-train Learning Rate & 1e-3                                            & 1e-3             & 1e-3            & 1e-3           & 1e-3              & 1e-2              \\
            Pre-train Epochs        & 200 (D $\to$ A) \& 500 (A $\to$ D)              & 500              & 100             & 500            & 500               & 200               \\
            Pre-train Weight Decay  & \multicolumn{6}{c}{0 for all datasets}                                                                                                        \\
            Fine-tune Learning Rate & 1e-3                                            & 1e-3             & 1e-3            & 1e-3           & -                 & -                 \\
            Fine-tune Epochs        & 3000                                            & 3000             & 3000            & 3000           & -                 & -                 \\
            Fine-tune Weight Decay  & \multicolumn{6}{c}{1e-5 for all datasets}                                                                                                     \\
            k (SP)                  & 2                                               & 2                & 1               & 1              & -                 & 1                 \\
            k (SP++)                & 3                                               & 3                & -               & 3              & 3                 & 10                \\
            repeat (SP++)           & 100                                             & 100              & -               & 50             & 50                & 10                \\
            Pooling                 & MEAN                                            & ATTN             & GCN             & GCN            & GCN               & MEAN              \\
            Optimizer               & \multicolumn{6}{c}{AdamW used for all datasets}                                                                                               \\
            Early Stop              & \multicolumn{6}{c}{200 for all datasets}                                                                                                      \\
            \bottomrule
        \end{tabular}
    }
    \caption{Hyper-parameters for our proposed Subgraph Pooling and Subgraph Pooling++. }
    \label{tab:hyperparameter}
\end{table*}

\section{Additional Experiments}

\subsection{Facebook Dataset}

\begin{figure*}[!h]
    \centering
    \subfloat{\includegraphics[width=0.8\linewidth]{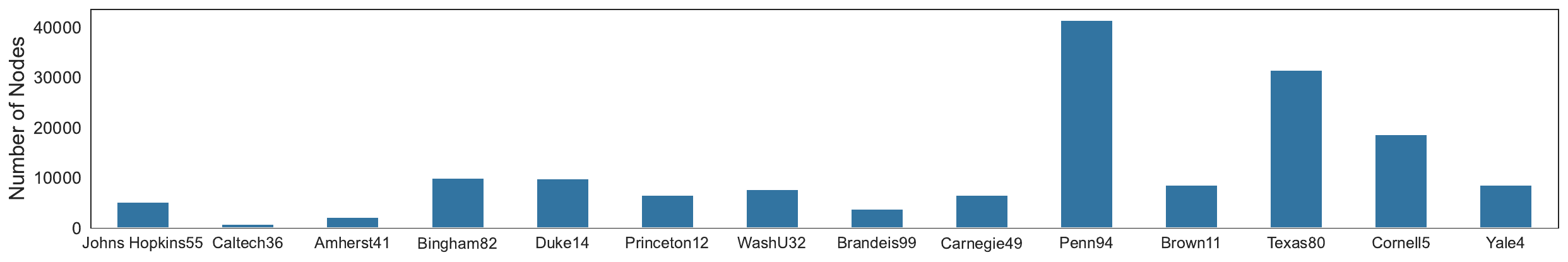}} \\
    \subfloat{\includegraphics[width=0.8\linewidth]{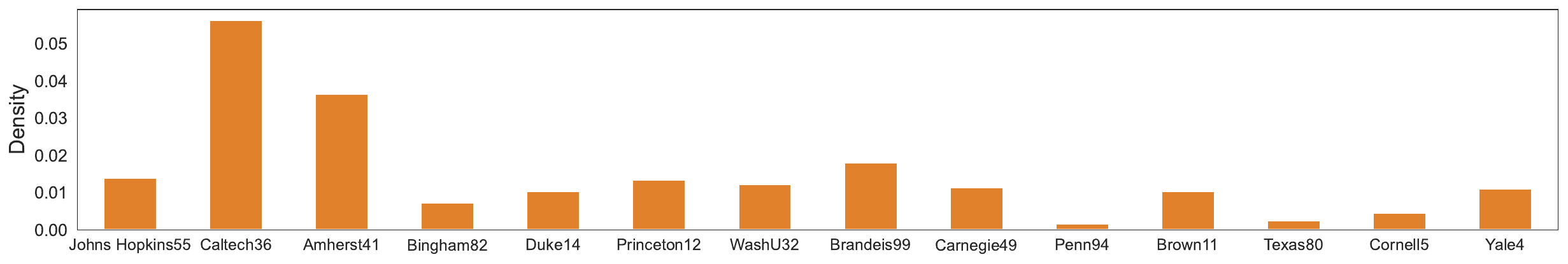}} \\
    \subfloat{\includegraphics[width=0.8\linewidth]{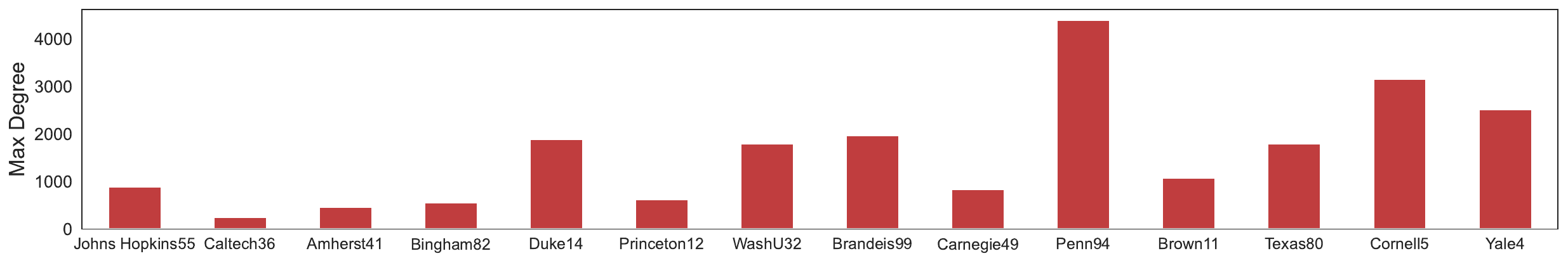}}
    \caption{Statistics of \texttt{Facebook} dataset.}
    \label{fig:statistics facebook}
\end{figure*}

We conduct experiments on \texttt{Facebook} dataset. This dataset contains 100 snapshots of Facebook friendship networks from 2005, with each network representing users from a specific American university. For our experiments, we selected 14 networks, including those from John Hopkins, Caltech, Amherst, Bingham, Duke, Princeton, WashU, Brandeis, Carnegie, Penn, Brown, Texas, Cornell, and Yale. The test datasets are Penn, Brown, and Texas, while Cornell and Yale are utilized for evaluation. The training sets are combinations of the remaining nine graphs. These datasets display unique sizes, densities, and degree distributions, as depicted in Figure \ref{fig:statistics facebook}. Notably, the distributions of the testing graphs differ significantly from those of the training and validation sets, providing a rigorous assessment of out-of-distribution (OOD) performance. Moreover, we use the No Fine-tune setting due to the absence of training data on the target graphs. The effectiveness of our methods is evident in Table \ref{tab:facebook}, where our GNN-SP outperformed ERM and EERM in 8 out of 9 settings. This result highlights the robust transferability of our approach across multiple sources with varied distributions and multiple target environments.

\begin{table*}[!h]
    \resizebox{\linewidth}{!}{
        \begin{tabular}{l|ccc|ccc|ccc}
            \toprule
                                             & \multicolumn{3}{c|}{Penn} & \multicolumn{3}{c|}{Brown} & \multicolumn{3}{c}{Texas}                                                                                                 \\ \cmidrule{2-4}\cmidrule{5-7}\cmidrule{8-10}
            Combinations                     & ERM                       & EERM                       & GNN-SP                    & ERM        & EERM       & GNN-SP              & ERM        & EERM       & GNN-SP              \\ \midrule
            John Hopkins + Caltech + Amherst & 50.48±1.09                & 50.64±0.25                 & \textbf{52.36±0.54}       & 54.53±3.93 & 56.73±0.23 & \textbf{56.86±0.04} & 53.23±4.49 & 55.57±0.75 & \textbf{55.93±1.00} \\
            Bingham + Duke + Princeton       & 50.17±0.65                & 50.67±0.79                 & \textbf{51.81±0.52}       & 50.43±4.58 & 52.76±3.40 & \textbf{56.34±0.05} & 50.19+5.81 & 53.82±4.88 & \textbf{56.34±0.05} \\
            WashU + Brandeis + Carnegie      & 50.83± 0.17               & \textbf{51.52±0.87}        & 50.85±0.05                & 54.61+4.75 & 55.15±3.22 & \textbf{56.96±0.02} & 56.25+0.13 & 56.12+0.42 & \textbf{56.32±0.05} \\ \bottomrule
        \end{tabular}
    }
    \caption{Node classification performance across \texttt{Facebook} networks with GCN backbone.}
    \label{tab:facebook}
\end{table*}

\subsection{More Analysis}

\noindent\textbf{Varying Training Ratio.} The transfer learning performance highly relies on the number of training instances available on the target. To evaluate the sensitivity to varying training sample sizes, we consider four training ratios $\{0.1\%, 0.5\%, 1\%, 10\%\}$, as shown in Table 2 in the main body. Our proposed GNN-SP outperforms all baselines. We observe that transfer learning models achieve better performance than the model trained from scratch (No Transfer) with extremely limited labels. This is because the knowledge gained from the source provides beneficial regularization during the fine-tuning process. However, when the training ratio reaches $q=10\%$, standard GNNs acquire sufficient knowledge to surpass other baselines, with the exception of our GNN-SP.

\end{document}